\documentclass{article}
\usepackage[a4paper, total={6in, 9.5in}]{geometry}
\usepackage[dvipsnames]{xcolor}
\usepackage{mathtools} \usepackage{multirow}
\usepackage{booktabs}
\usepackage{colortbl}
\usepackage{subfigure}
\usepackage[export]{adjustbox}
\usepackage{afterpage}
\usepackage{wrapfig}
\usepackage[ruled,linesnumbered]{algorithm2e} 
\usepackage{newtxtext}

\usepackage[frozencache,cachedir=.]{minted}

\usepackage{wrapfig}
\usepackage{tabularx}
\usepackage{makecell}

\usepackage{amsmath,amsfonts,bm}

\def\eqref#1{equation~\ref{#1}}

\def\1{\bm{1}}

\DeclareMathAlphabet{\mathsfit}{\encodingdefault}{\sfdefault}{m}{sl}
\SetMathAlphabet{\mathsfit}{bold}{\encodingdefault}{\sfdefault}{bx}{n}

\usepackage{xspace}
\newcolumntype{Y}{>{\centering\arraybackslash}X}
\newcommand{\ie}{{\emph{i.e.}},\xspace}

\newcommand{\eg}{{\emph{e.g.}},\xspace}

\newcommand{\etal}{{\emph{et al.}}}

\usepackage{amsthm}

\theoremstyle{remark}

\theoremstyle{definition}

\usepackage{soul}
\definecolor{almond}{rgb}{0.94, 0.87, 0.8}
\definecolor{antiquewhite}{rgb}{0.98, 0.92, 0.84}
\definecolor{electriclavender}{rgb}{0.96, 0.73, 1.0}
\definecolor{cloudgray}{rgb}{0.92, 0.92, 0.92}
\definecolor{dawnblue}{rgb}{0.84, 0.92, 1.0}
\definecolor{rdd}{rgb}{0.85, 0.8, 0.95}

\definecolor{cb2}{RGB}{146,   0,   0}
\definecolor{cb3}{RGB}{182, 109, 255}
\definecolor{cb1}{RGB}{ 0, 109, 219}
\definecolor{cb4}{RGB}{ 36, 255,  36}

\usepackage{color, colortbl}
\definecolor{Gray}{gray}{0.9}
\usepackage{hyperref}
 \hypersetup{
     colorlinks=true,
     linkcolor=MidnightBlue,
     filecolor=MidnightBlue,
     citecolor = MidnightBlue,
     urlcolor=MidnightBlue,
 }

\usepackage{url}

\usepackage{ifthen}
\usepackage{amssymb}
\newboolean{showcomments}
\setboolean{showcomments}{true} 
\ifthenelse{\boolean{showcomments}}
  {\newcommand{\nb}[2]{
    \fcolorbox{gray}{yellow}{\bfseries\sffamily\scriptsize#1}
    {\sf\small$\blacktriangleright$\textit{#2}$\blacktriangleleft$}
   }
   
  }
  {\newcommand{\nb}[2]{}
   
  }

\newcommand{\todonr}[1]{\smash{\fcolorbox{gray}{red}{?}}}

\title{SplitMixer: Fat Trimmed From MLP-like Models}

\author{
    Ali Borji$^1$,\; 
 Sikun Lin$^2\footnote{Authors contributed equally to this work.}$ \\
    $^1$Quintic AI, $^2$University of California, Santa Barbara \\
}

\def\eg{\emph{e.g.~}}

\def\etal{{\em et al.~}}
\def\ie{\emph{i.e.~}}

\begin{document}

\maketitle

\begin{abstract}
    We present SplitMixer, a simple and lightweight isotropic MLP-like architecture, for visual recognition. It contains two types of interleaving convolutional operations to mix information across spatial locations (spatial mixing) and channels (channel mixing).
    The first one includes sequentially applying two depthwise 1D kernels, instead of a 2D kernel, to mix spatial information. The second one is splitting the channels into overlapping or non-overlapping segments, with or without shared parameters, and applying our proposed channel mixing approaches or 3D convolution to mix channel information. Depending on design choices, a number of SplitMixer variants can be constructed to balance accuracy, the number of parameters, and speed. We show, both theoretically and experimentally, that SplitMixer performs on par with the state-of-the-art MLP-like models while having a significantly lower number of parameters and FLOPS. For example, without strong data augmentation and optimization, SplitMixer achieves around 94\% accuracy on CIFAR-10 with only 0.28M parameters, while ConvMixer achieves the same accuracy with about 0.6M parameters. The well-known MLP-Mixer achieves 85.45\% with 17.1M parameters. On CIFAR-100 dataset, SplitMixer achieves around 73\% accuracy, on par with ConvMixer, but with about 52\% fewer parameters and FLOPS. We hope that our results spark further research towards finding more efficient vision architectures and facilitate the development of MLP-like models. Code is available at \url{https://github.com/aliborji/splitmixer}.

\end{abstract}

\section{Introduction}
Architectures based exclusively on multi-layer perceptrons (MLPs)\footnote{The MLPs used in the MLP-Mixer, in particular those applied to the image patches, are in essence convolutions. But since each convolution is in fact an MLP, one can argue that MLP-Mixer only uses MLPs. The fact that matters is that parameters are shared across MLPs that are applied to image patches.}~\cite{mixer} have emerged as strong competitors to Vision Transformers (VIT)~\cite{vit} and Convolutional Neural Networks (CNNs)~\cite{krizhevsky2012imagenet,he2016deep}. They achieve compelling performance on several computer vision problems, in particular large-scale object classification. Further, they are very simple and efficient and perform on par with more complicated architectures. MLP-like models contain two types of layers to mix information across spatial locations (spatial mixing) and channels (channel mixing). These operations can be implemented via MLPs as in~\cite{mixer}, or convolutions as in~\cite{trockman}.

\begin{figure}
    \centering
    \includegraphics[width=7cm,height=5cm]{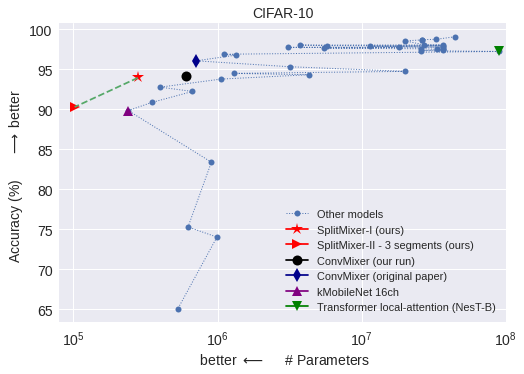}
    \includegraphics[width=7cm,height=5cm]{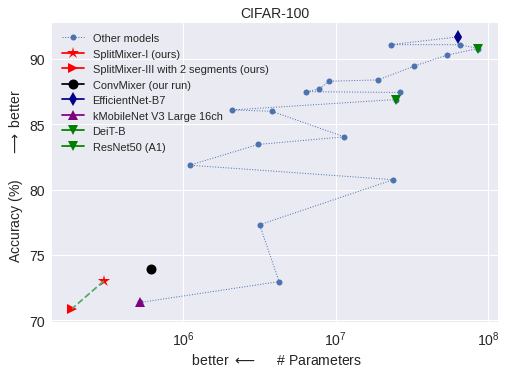}
    \vspace*{-1em}
    \caption{Comparison of our proposed SplitMixer architectures with state-of-the-art models that do not use external data for training. Results are shown over CIFAR-\{10,100\} datasets. Notice that {\bf we have not optimized our models for the best performance}. Rather, we ran the ConvMixer and our models using the exact same code, parameters, and machines to measure how much we can save parameters and computation relative to ConvMixer. Please consult~\cite{trockman} for a more detailed comparison of ConvMixer with other models. We have borrowed some data from \url{https://paperswithcode.com/} to generate these plots.}
    \label{fig:sota}
\end{figure}

We propose the SplitMixer, a conceptually and technically simple, yet very efficient architecture in terms of accuracy, the number of required parameters, and computation. Our model is similar in spirit to the ConvMixer~\cite{trockman} and MLP-Mixer~\cite{mixer} models\footnote{MLPMixer generalizes the ConvMixer.} in that it accepts image patches as input, dissociates spatial mixing from channel mixing, and maintains equal size and resolution throughout the network, hence an isotropic architecture. Similar to ConvMixer, it uses standard convolutions to achieve the mixing steps. Unlike ConvMixer, however, it uses 1D convolutions to mix spatial information. This modification maintains the accuracy but does not lower the number of parameters significantly. The biggest reduction in the number of parameters is achieved by how we modify channel mixing. Instead of applying $1\times1$ convolutions across all channels, we apply them to channel segments that may or may not overlap each other. We implement this part with our ad-hoc solutions or with 3D convolution. This way, we find some architectures that are very frugal in terms of model size and computational needs, and at the same time exhibit high accuracy. To illustrate the efficiency of the proposed modifications, in Figure~\ref{fig:sota} we plot accuracy \emph{vs.} number of parameters for our models and state-of-the-art models that do not use external data for training. Over CIFAR-\{10,100\} datasets, in the low-parameter regime, our models push the envelope towards the top-left corner, which means a better trade-off between accuracy and model size (also speed). Our models even outperform some very well-known architectures such as MobileNet~\cite{mobilenets}.

Despite its simplicity, SplitMixer achieves excellent performance. Without
strong data augmentations, it attains around 94\% Top-1 accuracy on CIFAR10 with only 0.27M parameters and 71M FLOPS. ConvMixer achieves the same accuracy but with 0.59M parameters and 152M FLOPS (almost twice more expensive). MLP-Mixer can only achieve 85.45\% with 17.1M parameters and 1.21G FLOPS. ResNet50~\cite{he2016deep} achieves 80.76\% using 23.84M parameters and, MobileNet~\cite{mobilenets} attains 89.81\% accuracy using 0.24M parameters.

\vspace{5pt}
In summary, our main contributions are as follows:
\begin{itemize}
    \item Applying 1D depthwise convolution sequentially across width and height for spatial mixing. We are inspired by the extensive use of spatial separable convolutions and depthwise separable convolutions in the literature (\eg ~\cite{mobilenets}),
    \item Splitting the channels into overlapping or non-overlapping segments and applying $1\times1$ pointwise convolution to segments for channel mixing,
    \item Theoretical analyses and empirical support for computational efficiency of the proposed solution, as well as measuring model throughput, and
    \item Ablation analyses to determine the contribution of different model components.
\end{itemize}

\section{SplitMixer}
\label{model}

\begin{figure}[t!]
    \centering
    \includegraphics[width=1\textwidth]{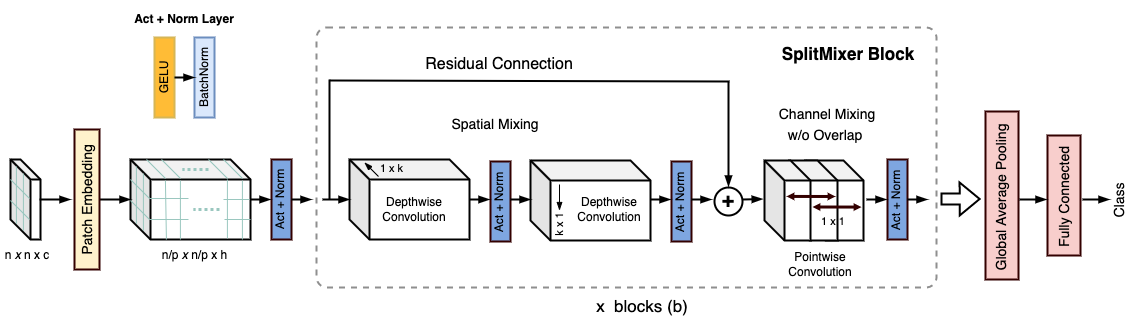}
    \vspace*{-2em}
    \caption{Basic architecture of SplitMixer. The input image is evenly divided into several image patches which are tokenized with linear projections. A number of 1D depthwise convolutions (spatial mixing) and pointwise convolutions (channel mixing) are repeatedly applied to the projections. For channel mixing, we split the channels into segments (hence the name SplitMixer) and perform convolution on them. We implement this part with our ad-hoc solutions or 3D convolution. Finally, a global average pooling layer followed by a fully-connected layer is used for class prediction.}
    \label{fig:SplitMixer}
\end{figure}

The overall architecture of SplitMixer is depicted in Figure~\ref{fig:SplitMixer}. It consists of a patch embedding layer followed by repeated applications of fully convolutional SplitMixer blocks. Patch embeddings with patch size $p$ and embedding dimension $h$ are implemented as 2D convolution with $c$ input channels (3 for RGB images), $h$ output channels, kernel size $p$, and stride $p$:

\begin{equation}
    z_0 = \mathsf{N}\left(\sigma\{\text{\texttt{Conv}}_{c \to h}(I, \text{\texttt{stride=}}p, \text{\texttt{kernel\_size}=}p)\}\right)
\end{equation}
\noindent where $\mathsf{N}$ is a normalization technique (\eg BatchNorm~\cite{ioffe2015batch}), $\sigma$ is an element-wise nonlinearity (\eg GELU~\cite{gelu}), and $\mathbf{I} \in \mathbb{R}^{n \times n \times c}$ is the input image. The SplitMixer block itself consists of two 1D depthwise convolutions (\ie grouped convolution with groups equal to the number of channels $h$) followed by several pointwise convolutions with kernel size $1\times1$. 
Each convolution is followed by nonlinearity and normalization \footnote{We use GELU and BatchNorm throughout the paper, except in ablation experiments.}. Therefore, each block can be written as:%
\begin{align}
    z_l' &= \mathsf{N}\left(\sigma\{\text{\texttt{ConvDepthwise}}(z_{l-1})\}\right) \ \ \ \ \ \ \ \ \ \ \ \ // \ \ \text{1}\times\text{k Conv across width}\\
    z_l' &= \mathsf{N}\left(\sigma\{\text{\texttt{ConvDepthwise}}(z_l')\}\right) + z_{l-1} \label{eq-depthwise} \ \ \ \ \ // \ \ \text{k}\times\text{1 Conv across height}\\ \ \
    z_l &= \mathsf{N}\left(\sigma\{\text{\texttt{ConvPointwise}}(z_l')\}\right) \label{eq-pointwise} \ \ \ \ \ \ \ \ \ \ \ \ \ \ \ \ // \ \ \text{1}\times\text{1 Conv across channels}
\end{align}
\looseness=-1
The SplitMixer block is applied $b$ times (indexed by \emph{l}), after which global pooling is applied to obtain a feature vector of size $h$. Finally, a softmax classifier maps this vector to the class label. Please see Appendix~\ref{appx:implementations} for PyTorch implementations. In what follows, we describe the spatial and channel mixing layers of the architecture.

\subsection{Spatial mixing}

We replace the $k \times k$ kernels\footnote{Throughout the paper, a tensor or a kernel is represented as \texttt{width} $\times$ \texttt{height} $\times$ \texttt{channels}.} in ConvMixer by two 1D kernels: 1) a $1 \times k$ kernel across width, and 2) a $k \times 1$ kernel across height. This reduces $k^2 \times h$ parameters to $2k \times h$ in each SplitMixer block. Similarly the $W \times H \times k^2$ FLOPS is reduced to $W \times H \times 2k$, where $W$ and $H$ are width and height of the input tensor $\mathbf{X} \in \mathbb{R}^{W \times H \times h}$, respectively. Therefore, separating the 2D kernel into two 1D kernels results in $\frac{k}{2}$ times savings in parameters and FLOPS. The two 1D convolutions are applied sequentially and each one is followed by a GELU activation and BatchNorm (denoted as ``Act + Norm'' in Figure~\ref{fig:SplitMixer}).

\subsection{Channel mixing}
\begin{wrapfigure}[15]{r}{0.45\textwidth}
 \vspace*{-5.3em}
  \begin{center}
      \includegraphics[width=0.32\textwidth]{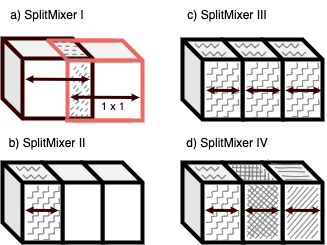}
  \end{center}
    \vspace*{-1.5em}
    \caption{\small{Channel mixing approaches: a) channels are split into two overlapping segments, and only one segment is convolved in each block (no parameter sharing across segments), b) channels are equally split into a number of segments, and only one segment is convolved in each block (no overlap or parameter sharing), c) all segments are convolved in each block and parameters are shared across segments, and d) all segments are convolved in each block (no parameter sharing).}}
  \label{fig:mixings}
  \vspace*{5.5em}
\end{wrapfigure}

We notice that most of the parameters in ConvMixer reside in the channel mixing layer. For $h$ channels and kernel size $k$ ($h >> k$), in each block there are $h \times k^2$ parameters in the spatial mixing part and $h^2$ parameters in the channel mixing part. Thus, the fraction of parameters in the two parts is $\frac{h \times k^2}{h^2} = \frac{k^2}{h}$ which is much smaller than 1 (\eg $5^2/256$). Therefore, most of the parameters are used for channel mixing.

\vspace{5pt}
\noindent {\bf Implementation using 3D convolution.} The basic idea here is to utilize 3D convolutions with certain strides. The output will be a set of interleaved maps coming from different segments. The same 3D kernel (with shared parameters) is applied to all segments. A certain number of 3D kernels will be needed to obtain an output tensor with the same number of channels as the input. Applying $m$ 3D kernels of size $1 \times 1 \times \frac{h}{m}$ and stride $\frac{h}{m}$ (assume $h$ is divisible by $m$), will require $\frac{h^2}{m}$ parameters. Hence, more parameters and computation will be saved by increasing the number of segments (\ie smaller 3D kernels). While being easy to implement, using 3D convolution has some restrictions. For example, kernel parameters have to be shared across segments, and all segments have to be convolved. Further, we find that channel mixing using 3D convolution is much slower than our other approaches (mentioned next). Please see Appendix~\ref{appx:3dConv} for the implementation of the channel mixing layer using 3D convolution.

\vspace{5pt}
\noindent {\bf Other channel mixing approaches.} A number of approaches are proposed that differ depending on whether they allow overlap or parameter sharing among segments. They offer different degrees of trade-off in accuracy, number of parameters, and FLOPS. 
Notice that both of our spatial and channel mixing modifications can be used in tandem or separately. In other words, they are independent of each other. The channel mixing approaches are shown in Figure~\ref{fig:mixings} and are explained below.

\subsubsection{SplitMixer-I: Overlapping segments, no parameter sharing, update one segment per block} 
The input tensor is split into two overlapping segments along the channel dimension. The intuition here is that the overlapped channels allow efficient propagation of information from the first segment to the second. Let $m$ be the size of each segment and a fraction of $h$, \ie $m = \alpha \times h, \alpha > 0.5$. The two segments can be represented as $X[:m]$ and $X[h-m:]$ in PyTorch. For instance, for $\alpha=2/3$, one-third of the middle channels are shared between the two segments. We choose to apply convolution to only one segment in each block, \eg the left segment in odd blocks and the right segment in even blocks. $m$ number of $1 \times 1$ convolutions are applied to the segment that should be updated. Therefore, the output has the same number of channels as the original segment, which is then concatenated to the other (unaltered) segment. The final output is a tensor with $h$ channels to be processed in the next block. In the experiments, we choose $\alpha = \frac{i}{2i-1}, i \in \{2 \cdots 6\}$. The reduction in parameters per block can be approximated as\footnote{For simplicity, here we discard bias, BatchNorm, and optimizer parameters.}:
\begin{equation}
    h^2 - (\alpha \times h)^2 = (1 - \alpha^2)  \times h^2 
\end{equation} 
\noindent which means $1 - \alpha^2$ fraction of parameters are reduced (\eg 56\% parameter reduction for $\alpha=2/3$). Notice that the bigger the $\alpha$, the less saving in parameters. Similarly, the reduction in FLOPS can be approximated as:
\begin{equation}
    W \times H \times h \times h   -   W \times H \times (\alpha \times h) \times (\alpha \times h) = 
    (1 - \alpha^2) \times W \times H \times h^2 
\end{equation} 
These equations show the saving in FLOPS is the same as the saving in parameters. We also tried a variation of this design which is updating both segments in the same block. This new variation has fewer parameters and FLOPS than ConvMixer. It saves less parameters compared to SplitMixers (ratio equal to $1 - 2\alpha^2$) but achieves slightly higher accuracy (\ie trade-off in favor of accuracy). Results are shown in Appendix~\ref{appx:SplitMixerV}.

\subsubsection{SplitMixer-II: Non-overlapping segments, no parameter sharing, update one segment per block} 
We first split the $h$ channels into $s$ non-overlapping segments, each with size $\frac{h}{s}$, along the channel dimension\footnote{Notice that when $h$ is not divisible by the number of segments, the last segment will be longer (\eg dividing $h=256$ into 3 segments means the segments would have dimension [85, 85, 86], in order).}. In each block, only one segment is convolved and updated. Parameters are not shared across the segments. 
Following the above calculation, saving in parameters and FLOPS is $1 - \frac{1}{s^2}$. For example, for $s=2$, roughly 75\% of the parameters are reduced. The same argument holds for FLOPS.

\subsubsection{SplitMixer-III: Non-overlapping segments, parameter sharing, update all segments per block} 
Here, $h$ channels are split into $s$ non-overlapping segments with shared parameters. Notice that under this setting, $h$ must be divisible by $s$ in order to get all the channels convolved. All segments are convolved and updated simultaneously in each block. Due to parameter sharing, the reduction in parameters is the same as SplitMixer-II, \ie the two SplitMixers have the same number of parameters for the same number of segments. The number of FLOPS, however, is higher now since computation is done over all $s$ segments. The number of FLOPS is the same as SplitMixer-IV, which will be calculated in the following subsection.

\subsubsection{ SplitMixer-IV: Non-overlapping segments, no parameter sharing, update all segments per block}  
This approach is similar to SplitMixer-III with the difference that here parameters are not shared across the segments. All segments are convolved and updated, and the results are concatenated. The reduction in parameters per block is:
\begin{equation}
        h^2 - s \times (h/s)^2  = (1 - \frac{1}{s}) \times h^2 
\end{equation} 
\noindent which results in $1 - \frac{1}{s}$ parameter saving. For example, 66.6\% of the parameters are reduced for $s=3$. More savings can be achieved with more segments. The reduction in FLOPS is:
\begin{equation}
    W \times H \times h \times h   -  s \times (W \times H \times \frac{h}{s} \times \frac{h}{s}) = 
    (1 - \frac{1}{s}) \times W \times H \times h^2 
\end{equation} 

\noindent which means the same saving in FLOPS as in parameters. 

\vspace{5pt}
\noindent {\bf Comparison of channel mixing approaches.} 
Among the mixing approaches, the SplitMixer-II saves the most parameters and computation but achieves lower accuracy. SplitMixer-I strikes a good balance between accuracy and model size (and FLOPS) thanks to its partial channel sharing. We assumed the same number of blocks in all mixing approaches. In practice, a smaller number of blocks might be required when all segments are updated simultaneously in each block. Notice that apart from these approaches, there may be some other ways to perform channel mixing. For example, in SplitMixer-I, parameters can be shared across the overlapped segments, or multiple segments can overlap. We leave these explorations to future research (See Appendix~\ref{appx:SplitMixerV}). We have also empirically measured the amount of potential saving in parameters and FLOPS over CIFAR-10 and ImageNet datasets, for model specifications mentioned in the next section. Results are shown in Appendix~\ref{appx:params}.

\vspace{5pt}
\noindent {\bf Naming convention.} We name SplitMixers after their hidden dimension $h$ and number of blocks $b$ like SplitMixer-A-h/b, where A is a specific model type (I, II, $\ldots$).

\section{Experiments and Results}
\label{sec-experiments}

We conducted several experiments to evaluate the performance of SplitMixer in terms of accuracy, the number of parameters, and FLOPS. 
Our goal was not to obtain the best possible accuracy. Rather, we were interested in knowing whether and how much parameters and computation can be reduced relative to ConvMixer. To this end, we used the exact same code, parameters, and machines to run the models. A thorough comparison of ConvMixer with other models is made in \cite{trockman}. We used PyTorch to implement our model and a Tesla V100 GPU with 32GB RAM to run it.

\subsection{Experimental setup} 

We used RandAugment~\cite{randaug}, random horizontal flip, and gradient clipping. 
Due to limited computational resources, \emph{we did no perform extensive hyperparameter tuning}, so better results than those reported here may be possible\footnote{Unlike ConvMixer, we are not using the \texttt{timm} framework. This framework offers a wider range of data augmentations, such as Mixup and Cutmix, which are not provided by PyTorch (torchvision.transforms). We expect better results using timm.}. All models were trained for 100 epochs with batch size 512 over CIFAR-\{10,100\} and 64 over Flowers102 and Food101 datasets. Across all datasets, $h$ and $b$ were set to 256 and 8.  

We used AdamW~\cite{adamw} as the optimizer, with weight decay set to 0.005 (0.1 for Flowers102). The learning rate (lr) was adjusted with the OneCycleLR scheduler (max-lr was set to 0.05 for CIFAR-\{10,100\}, 0.03 for Flowers102, and 0.01 for Food101). We utilized the ithop library\footnote{https://github.com/Lyken17/pytorch-OpCounter} for measuring the number of parameters and FLOPS.

\subsection{Results on CIFAR-\{10,100\} dasasets}
\label{apx-cifar}

Both datasets contain 50,000 training images and 10,000 test images (resolution is 32 $\times$ 32); each class has the same number of samples. We set $p=2$ and $k=5$ over both datasets. 

\begin{figure}[htbp]
    \centering
    \includegraphics[width=.45\textwidth]{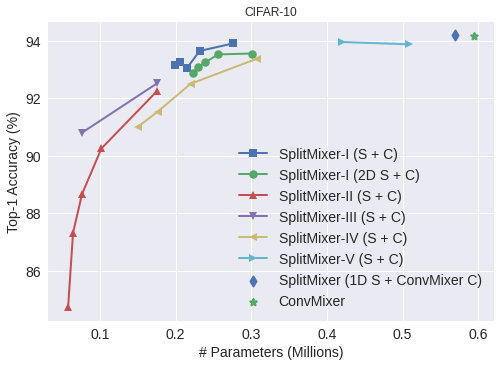}
    \includegraphics[width=.45\textwidth]{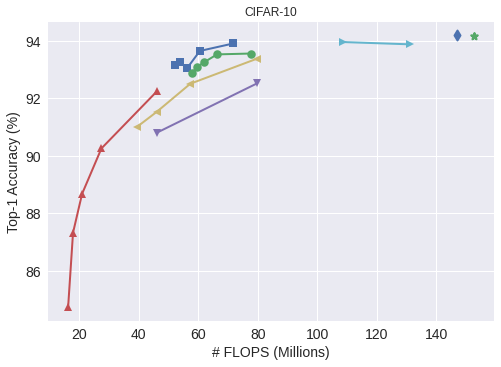}

    \includegraphics[width=.45\textwidth]{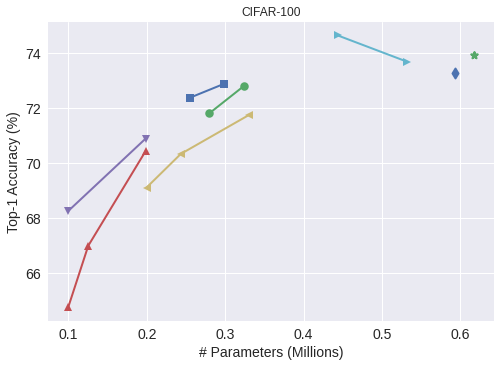}
    \includegraphics[width=.45\textwidth]{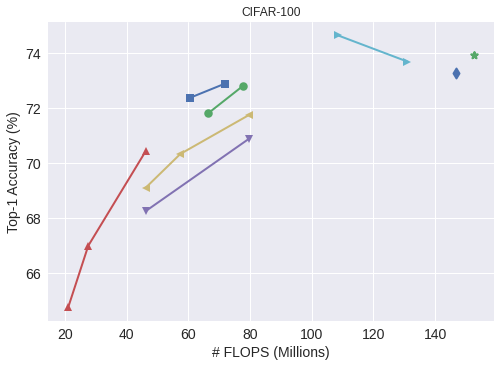}
    
    \includegraphics[width=.45\textwidth]{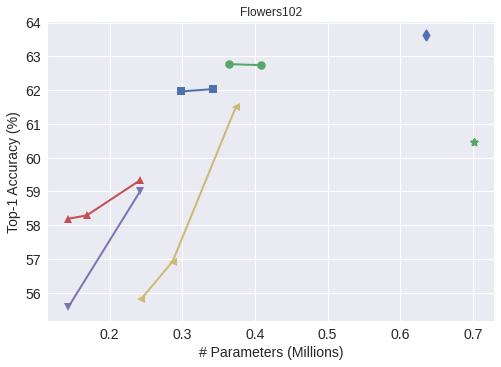}
    \includegraphics[width=.45\textwidth]{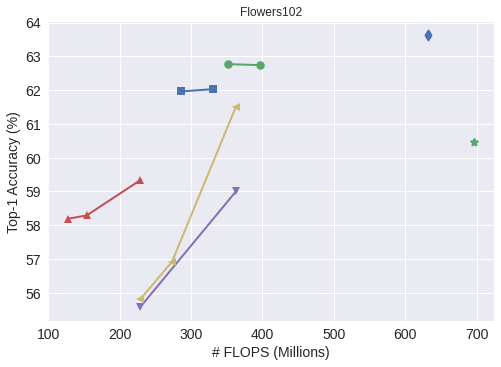}
    
    \includegraphics[width=.45\textwidth]{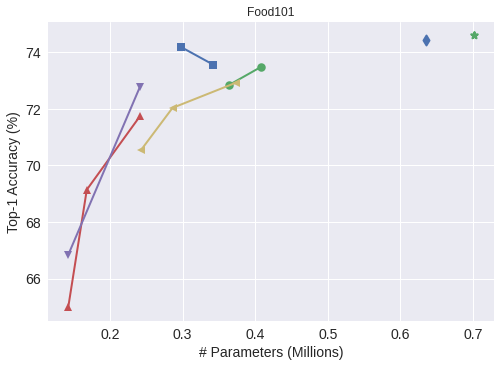}
    \includegraphics[width=.45\textwidth]{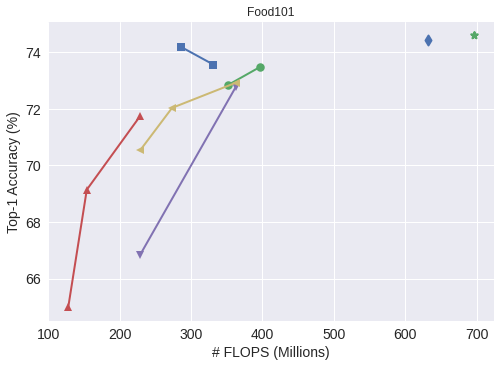}
    
    \vspace*{-.5em}
    \caption{Performance of SplitMixer variants. Left) Accuracy \emph{vs.} parameters, Right) Accuracy \emph{vs.} FLOPS. S stands for 1D spatial convolution and C stands for $1 \times 1$ pointwise convolution over channel segments. We plug in our components into ConvMixer, denoted here as ``2D + C'' (2D convolution kernels plus our channel mixing approach) and ``1D S + ConvMixer C'' (our 1D kernels plus channel mixing as is done in ConvMixer, \ie $1 \times 1$ convolution across all channels without splitting). Data points are for different values of split ratio or number of segments depending on the model type. We have collected more data points on CIFAR-10 than other datasets. Notice that the ratio of FLOPS over the number of parameters is almost the same for all models except SplitMixer-III, where this ratio is higher since all segments are updated in each block and parameters are shared across segments (see Fig.~\ref{fig:mixings}). That is why the plots for parameters and FLOPS are almost the same for each model, except SplitMixer-III. Please see Appendix~\ref{appx:SplitMixerV} for a description of SplitMixer-V.}
    \label{fig:res}
\end{figure}

Shown in the top panel of Figure~\ref{fig:res}, ConvMixer scores slightly above 94\% on CIFAR-10\footnote{The original ConvMixer paper~\cite{trockman} has reported 96\% accuracy on CIFAR-10 with Mixup and Cutmix data augmentation with 0.7M parameters. We, thus, expect even better results for SplitMixer with stronger data augmentation.}. 
SplitMixer-I has about the same accuracy as ConvMixer but with less than 0.3M parameters which are almost half of the ConvMixer parameters. The same statement holds for FLOPS as shown in the right panels of Figure~\ref{fig:res}. SplitMixer with 1D spatial mixing and regular $1\times1$ channel mixing as in ConvMixer (denoted as ``SplitMixer 1D S + ConvMixer C'' in the Figure) attains about the same accuracy as ConvMixer with slightly lower parameters and FLOPS. SplitMixer-I with 2D spatial kernels and segmented channel mixing (denoted as ``SplitMixer 2D S + C'') performs on par with SplitMixer-I. Performance of the SplitMixer-II quickly drops with more segments (and subsequently fewer parameters). SplitMixers III and IV also perform well (above 90\%). Interestingly, with only about 76K parameters, SplitMixer-III reaches about 91\% accuracy. SplitMixer-V (Appendix~\ref{appx:SplitMixerV}), performs close to ConvMixer, but it does not save many parameters or FLOPS.

Qualitatively similar results are obtained over the CIFAR-100 dataset. Here, ConvMixer scores 73.9\% accuracy, above the 72.5\% by SplitMixer-I, but with twice more parameters and FLOPS. 

On both datasets, increasing the number of segments saves more parameters and FLOPS but at the expense of accuracy. Interestingly, SplitMixer-I with only channel mixing does very well. Our channel mixing approaches are much more effective than 1D spatial mixing in terms of lowering the number of parameters and FLOPS. 

\vspace{5pt}
\noindent {\bf Results using 3D convolution.}
We experimented with a model that uses $128$ kernels of size $1 \times 1 \times 128$ and stride $128$ along the channel dimension (\ie channels are partitioned into two non-overlapping segments each of size $128$) for channel mixing. This model scores 93.09\% and 71.99\% on CIFAR-10 and CIFAR-100, respectively. While having similar accuracy, the number of parameters (about 0.2 M), and FLOPS (about 0.08 G) as SplitMixer models, this model is much slower to train (each epoch takes twice more time). Notice that, among the channel mixing approaches, only SplitMixer-III can be considered as 3D convolution. Thus, performing channel mixing through strided 3D convolution is a subset of our proposed solutions.

\subsection{Results on Flowers102 and Food101 datasets}
\label{apx-flowers-foods}

Flowers102 contains 1020 training images (10 per class) and 6149 test images. Food101 contains 750 training images and 250 test images for each of its 101 classes. We used larger patch ($p=7$) and kernel sizes ($k=7$) since image size is bigger in these datasets (both resized to 224 $\times$ 224). 

Results are shown in Figures~\ref{fig:res}. The patterns are consistent with what we observed over CIFAR datasets. SplitMixer variants, with small number of segments, perform close to the ConvMixer. Over the Flowers102 dataset, SplitMixer-I scores 62.03\%, higher than the 60.47\% by ConvMixer. Similarly, over Food101, SplitMixer-I scores 1\% lower than ConvMixer, but with less than half of ConvMixer's parameters and FLOPS. In general, increasing the overlap between segments (by raising $\alpha$ in SplitMixer-I) or reducing the number of segments enhances the accuracy but also increases the number of parameters across datasets (not conclusive on Food101 dataset). Further, SplitMixer is effective over both small and large datasets.

\subsection{Comparison with state of the art}
Table~\ref{tab:res} shows a comparison of SplitMixer with models from MLP, Transformer, and CNN families. Some results are borrowed from~\cite{lv2022mdmlp} where they trained the models for 200 epochs, whereas here we trained our models for 100 epochs. While the experimental conditions in~\cite{lv2022mdmlp} might not be exactly the same as ours, cross-examination still provides insights into how our models fare compared to others, in particular the MLP-based models. We have also provided additional comparative results in Figure~\ref{fig:sota}. Our models outperform other models while having significantly smaller sizes and computational needs. 
For example, SplitMixer-I has about the same number of parameters as ResNet20, but is about 2\% better on CIFAR-10 and 5\% better on CIFAR-100. Over Flowers102, SplitMixer drastically outperforms other models in all three aspects, including accuracy, number of parameters, and FLOPS. Both SplitMixer and ConvMixer are on par with other models on the Food101 dataset, with ConvMixer performing slightly better.

\begin{table}
\centering
    \centering
    \begin{tabular}{*{8}{c}}
        \toprule
        
        \multirow{2}{*}{Model Family} & \multirow{2}{*}{Model} & \multirow{2}{*}{Params (M)} & \multirow{2}{*}{FLOPS (G)} & \multicolumn{2}{c}{Top-1 ACC (\%)} \\ \cmidrule(lr){5-6}
        & & & & \small{CIFAR10} & \small{CIFAR100} \\ \midrule
        
        CNN & ResNet20~\cite{he2016deep} & \textbf{0.27} & \textbf{0.04} & 91.99 & 67.39 \\
        Transformer & ViT~\cite{vit} & 2.69 & 0.19 & 86.57 & 60.43 \\ \cmidrule(lr){1-6}
        MLP & AS-MLP~\cite{lian2021mlp} & 26.20 & 0.33 & 87.30 & 65.16 \\
        '' & gMLP~\cite{liu2021pay} & 4.61 & 0.34 & 86.79 & 61.60 \\
        '' & ResMLP~\cite{resmlp} & 14.30 & 0.93 & 86.52 & 61.40 \\
        '' & ViP~\cite{permutator} & 29.30 & 1.17 & 88.97 & 70.51 \\
        '' & MLP-Mixer~\cite{mixer} & 17.10 & 1.21 & 85.45 & 55.06 \\ 
        '' & \small{S-FC ($\beta$-LASSO)}~\cite{neyshabur2020towards} & - & - & 85.19 & 59.56 \\ 
        '' & MDMLP~\cite{lv2022mdmlp} \label{mdmlp1} & 0.30 & 0.28 & 90.90 & 64.22 \\ \cmidrule(lr){1-6}
        '' & ConvMixer~\cite{trockman2022patches} & 0.60 & 0.15 & \textbf{94.17} & \textbf{73.92} \\ \cmidrule(lr){1-6} 
        '' & SplitMixer-I (ours) & 0.28 & 0.07 & 93.91 & 72.44 \\ 
        
        \bottomrule
    \end{tabular}

    \medskip{}\medskip{}\medskip{}

    \begin{tabular}{*{8}{c}}
        \toprule
        \multirow{2}{*}{Model Family} & \multirow{2}{*}{Model} & \multirow{2}{*}{Params (M)} & \multirow{2}{*}{FLOPS (G)} & \multicolumn{2}{c}{ACC (\%)} \\ \cmidrule(lr){5-6}
        & & & & \small{Flowers102} & \small{Food101} \\ \midrule
        
        CNN & ResNet20~\cite{he2016deep} & \textbf{0.28} & 2.03 & 57.94 & 74.91 \\
        Transformer & ViT~\cite{vit} & 2.85 & 0.94 & 50.69 & 66.41 \\ \cmidrule(lr){1-6}
        MLP & AS-MLP~\cite{lian2021mlp} & 26.30 & 1.33 & 48.92 & 74.92 \\
        '' & gMLP~\cite{liu2021pay} & 6.54 & 1.93 & 47.35 & 73.56 \\
        '' & ResMLP~\cite{resmlp} & 14.99 & 1.23 & 45.00 & 68.40 \\
        '' & ViP~\cite{permutator} & 30.22 & 1.76 & 42.16 & 69.91 \\
        '' & MLP-Mixer~\cite{mixer} & 18.20 & 4.92 & 49.41 & 61.86 \\ 
        '' & MDMLP~\cite{neyshabur2020towards} & 0.41 & 1.59 & 60.39 & \textbf{77.85} \\ \cmidrule(lr){1-6}
        '' & ConvMixer~\cite{trockman2022patches} & 0.70 & 0.70 & 60.47 &  74.59\\ \cmidrule(lr){1-6} 
        '' & SplitMixer-I (ours) & 0.34 & \textbf{0.33} & \textbf{62.03} & 73.56 \\ 
        
        \bottomrule
    \end{tabular}
    \caption{Comparison with other models. The best numbers in each column are highlighted in bold. The number of parameters and FLOPS are averaged over CIFAR-10 and CIFAR-100 for our models. Notice that some variants of SplitMixer perform better than the numbers reported here over Flowers102 and Food101 datasets. Results, except ConvMixer and our model, are borrowed from~\cite{lv2022mdmlp} where they have trained models for 200 epochs. We have trained ConvMixer and SplitMixer for 100 epochs. } \label{tab:res}       

\end{table}

\subsection{Ablation experiments}

We conducted a series of ablation experiments to study the role of different design choices and model components. Results are shown in Table~\ref{tab:ablate} over CIFAR-\{10,100\} datasets. We took the SplitMixer-I as the baseline and discarded or added pieces to it. Our findings are summarized below:

\begin{table}[h]
\begin{center}
\begin{tabularx}{0.65\textwidth}{l|Y|Y}
\toprule
\multicolumn{3}{c}{Ablation of SplitMixer-I-256/8 on CIFAR-$\{10,100\}$} \\
\midrule
\multicolumn{1}{c|}{Ablation} & \thead{CIFAR-10\\Acc. (\%)} & \thead{CIFAR-100\\Acc. (\%)}  \\
\midrule
SplitMixer-I (baseline) & 93.91   &  72.88 \\
\midrule
\ \ \ \ \ \ \ \ -- Residual in Eq.~\ref{eq-depthwise} & 92.24    &  71.34 \\ 
\ \ \ \ \ \ \ \ + Residual in Eq.~\ref{eq-pointwise} &  92.35    &  70.44 \\ 
\midrule
\ \ \ \ \ \ \ \ BatchNorm $\to$ LayerNorm &   88.28   &  66.60 \\
\ \ \ \ \ \ \ \ GELU $\to$ ReLU & 93.39     & 72.56  \\
\midrule
\ \ \ \ \ \ \ \ -- RandAug &  90.87    &  66.54 \\
\ \ \ \ \ \ \ \ -- Gradient Norm Clipping & 93.38     &   71.95\\

\midrule
SplitMixer-I (Spatial only) &   76.24   &  53.25 \\
SplitMixer-I (Channel only) &   64.21   &  40.46 \\

\midrule
One segment with size $\alpha \times h; \alpha=\frac{2}{3}$  & 76.28  &  51.28 \\

\bottomrule
\end{tabularx}
\caption{Ablation study of SplitMixer-I-256/8 with split ratio of 2/3.}
\label{tab:ablate}
\end{center}
\end{table}

\begin{itemize}
    \item Completely removing the residual connections does not hurt the performance much. These connections, however, might be important for very deep SplitMixers.
    \item Moving the residual connection to after channel mixing seems to hurt the performance. We find that the best place for the residual connections is right after spatial mixing.
    \item Switching to LayerNorm, instead of BatchNorm, leads to drastic performance drop.
    \item  The choice of activation function, GELU \emph{vs.} ReLU, is not very important. In fact, we found that using ReLU sometimes helps. 
    \item  Gradient norm clipping hinders the performance slightly, thus it is not very important.
    \item  Data augmentation, here as RandAug\footnote{Unlike ConvMixer, we do not have Mixup and CutMix}, is critical to gaining high performance.
    \item SplitMixer-I with only 1D spatial mixing, and no channel mixing, performs very poorly. The same is true for ablating the spatial mixing \ie having only channel mixing. We find that spatial mixing is more important than channel mixing in our models.
    \item Keeping only one of the segments in channel mixing, hence 1D spatial mixing plus channel mixing using $m$ channels ($m < h$), lowers the accuracy by a large margin. This indicates that there is a substantial benefit in having a larger $h$ and splitting it into segments (and having overlaps between them). Notice that in each block, only one segment is updated. In other words, simply lowering the number of channels does not lead to the gains that we achieve with our models. Any ConvMixer, small or large, can be optimized using our techniques.
\end{itemize}

\begin{figure}[t]
    \centering
    \includegraphics[width=5cm,height=7cm]{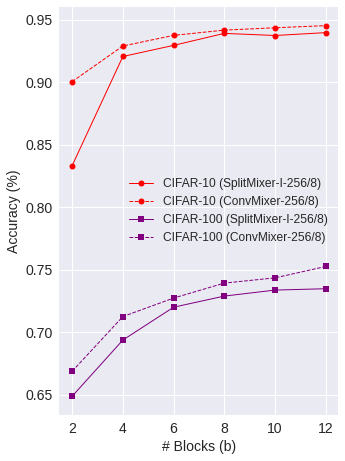}
    \includegraphics[width=5cm,height=7cm]{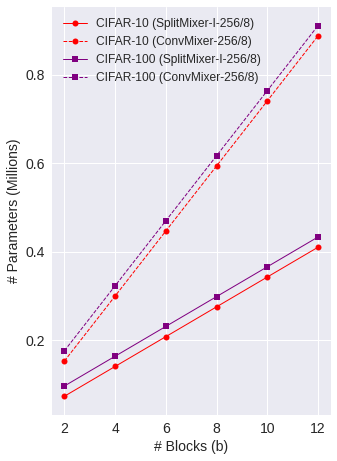}
    \includegraphics[width=5cm,height=7cm]{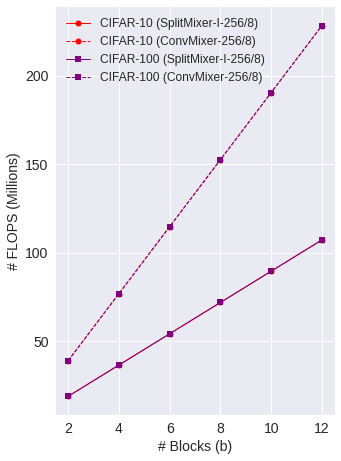}

    \caption{The role of the number of blocks $b$ on model performance. The FLOPS of models over CIFAR-100 are just slightly higher than CIFAR-10, thus not visible in the rightmost panel.} 
    \label{fig:blocks}
\end{figure}

\subsection{The role of the number of blocks}
We wondered about the utility of the proposed modifications over deeper networks. To this end, we varied the number of blocks $b$ of ConvMixer and SplitMixer-I in the range 2 to 10 in steps of 2, and trained the models. Other parameters were kept the same as above. As the results in Figure~\ref{fig:blocks} show, increasing the number of blocks improves the accuracy of both models on CIFAR\{10,100\}. SplitMixer-I performs slightly below the ConvMixer, but it has a huge advantage in terms of the number of parameters and FLOPS, in particular over deeper networks. The model size and computation grow slower for SplitMixer compared to ConvMixer.

\subsection{Model throughput}
We measured throughput using batches of 64 images on a single Tesla v100 GPU with 32GB RAM~\cite{nvidia2017nvidia}, averaging over 100 such batches. Similar to ConvMixer, we considered CUDA execution time rather than ``wall-clock'' time. Here, we used the network built for the FLOWER102 classification ($h=256$, $d=8$, $p=7$, $k=7$, and image size $224 \times 224$). We measured throughput when our model was the only process running on the GPU. Results are shown in Table~\ref{tab:throughput}. The throughput of our model is almost three times higher than ConvMixer. As expected, the throughput is higher with more segments in the channels because the number of FLOPS is lower. We also measured throughput using a Tesla v100 GPU with 16GB RAM. Results are presented in Appendix~\ref{appx:throughput}.

\begin{table}[h]
\begin{small}
\begin{center}
\begin{tabularx}{.8\textwidth}{l|YYYYYYY}
\toprule
Network& \multicolumn{7}{c}{Throughput \ \ (img/sec)} \\
\hline \hline
ConvMixer & \multicolumn{7}{c}{815.84} \\ \hline 
 & \multicolumn{7}{c}{Overlap ratio} \\ \cline{2-8}
         & 2/3 & 3/5 & 4/7 & 5/9 & 6/11 & - & -  \\ \cline{2-8}
SplitMixer-I       & 2097.55 & 2208.40 & 2210.06 & 2220.09 & 2231.42 & - &  -  \\ \hline 
        
 & \multicolumn{7}{c}{Number of segments} \\ \cline{2-8} 
                    & 2 & 3 & 4 & 5 & 6 & 7 & 8 \\ \cline{2-8}
SplitMixer-II       & 2322.02 & 2291.44 & 2440.16 & 2464.33 & 2474.318 & - & -  \\

SplitMixer-III       & 2112.290 & - &  2171.70 & - & - & - & 2185.61  \\

SplitMixer-IV      & 2110.92 & 2084.55 & 2170.57 & 2146.76 & - & - & - \\
\bottomrule
\end{tabularx}
\end{center}
\caption{Model throughput for SplitMixer-A-256/8 on a Tesla v100 GPU with 32GB RAM over a batch of 64 images of size 224 $\times$ 224, averaged over 100 such batches.}
\label{tab:throughput}
\end{small}

\end{table}

\subsection{Feature visualization}
In Appendix~\ref{appx:maps}, we plot the learned feature maps at successive layers of SplitMixer-I-256/8 model trained over CIFAR-10 with $\alpha=2/3$, $p=2$, and $k=5$. Notice that in each block, 170 channels are convolved (the first 170 channels in even blocks and the second 170 channels in odd blocks, with overlap). Some maps seem to be the output of convolution with oriented edge filters. Few others seem to be dead with no activation. We also found similar maps for the ConvMixer model (not shown).

\subsection{Negative results}
Here, we report some configurations that we tried but did not work. Initially, we attempted to use a number of fully connected layers to perform spatial and channel mixing, hoping the network would implicitly learn how to mix information. This model did not converge, indicating that perhaps inductive biases pertaining to spatial and channel mixing matter in these architectures. Also, sequentially applying $1\times1$ convolutions across the width, height and channels, with shared parameters across each, did not result in good accuracy, although some works (\eg~\cite{permutator}) have been able to make similar, but more complicated, designs to work.

\section{Related Work}

For about a decade, CNNs have been the de-facto standard in computer vision~\cite{he2016deep}. Recently, the Vision Transformers (ViT)~\cite{vit} by Dosovitskiy~\etal and its variants~\cite{cait,d2021convit,swin,touvron2021training,bao2021beit,wang2021pyramid}, and the multi-layer perceptron mixer (MLP-Mixer) by Tolstikhin~\etal~\cite{mixer} and its variants~\cite{resmlp,li2021convmlp} have challenged CNNs. These models have shown impressive results, even better than CNNs, in large-scale image classification. Unlike CNNs that exploit local convolutions to encode spatial information, vision transformers take advantage of the self-attention mechanism to capture global information. MLP-based models, on the other hand, capture global information through a series of spatial and channel mixing operations. 

MLP-Mixer borrows some design choices from recent transformer-based architectures~\cite{vaswani2017attention}. Following ViT, it converts an image to a set of patches and linearly embeds them to a set of tokens. These tokens are processed by a number of ``isotropic'' blocks, which are in essence similar to the repeated transformer-encoder blocks~\cite{vaswani2017attention}. 
For example, MLP-Mixer~\cite{mixer} replaces self-attention with MLPs applied across
different dimensions (\ie spatial and channel location mixing).
ResMLP~\cite{resmlp} is a data-efficient variation on this scheme.
CycleMLP~\cite{chen2021cyclemlp}, gMLP~\cite{liu2021pay}, and vision permutator~\cite{permutator},
conduct different approaches to perform spatial and channel mixing. For example, vision permutator~\cite{permutator} permutes a tensor along the height, width, and channel to apply MLPs. Some works attempt to bridge convolutional networks and vision transformers and use one to improve the other~\cite{cordonnier2019relationship,d2021convit,dai2021coatnet, guo2021cmt,wang2021pyramid,bello2019attention, ramachandran2019stand, bello2021lambdanetworks}.

We are primarily inspired by the ConvMixer~\cite{trockman}. This model introduces a simpler version of MLP-Mixer but is essentially the same. It replaces the MLPs in MLP-Mixer with convolutions. 
In general, Convolution-based MLP models are smaller than their heavy Transformer-, CNN-, and MLP-based counterparts.
Here, we show that it is possible to trim these models even more. Perhaps the biggest advantage of the MLP-based models is that they are easy to understand and implement, which in turn helps replicate results and compare models. To this end, we share our code publicly (\url{https://github.com/aliborji/splitmixer}).

\section{Discussion and Conclusion}

We proposed SplitMixer, an extremely simple yet very efficient model, that is similar in spirit to ConvMixer, ViT, and MLP-Mixer models. SplitMixer uses 1D convolutions for spatial mixing and splits the channels into several segments for channel mixing. It performs $1 \times 1$ convolution on each segment. Our experiments, even without extensive hyperparameter tuning, demonstrate that these modifications result in models that are very efficient in terms of the number of parameters and computation. In terms of accuracy, they outperform several MLP-based models and some other model types with similar size constraints. 
{\bf Our main point is that SplitMixer allows sacrificing a small amount of accuracy to achieve big gains in reducing parameters and FLOPS.}

The proposed solution based on separable filters, depthwise convolution, and channel splitting is quite efficient in terms of parameters and computation. However, if a network is already small, reducing the parameters too much may cause the network not to learn properly during training. Thus, a balance is required to enhance efficiency without significantly reducing effectiveness.

\vspace{5pt}
\noindent We propose the following directions for future research in this area: 

\begin{itemize}
    \item We suggest trying a wider range of hyperparameters and design choices for SplitMixer, such as strong data augmentation (\eg Mixup, Cutmix), deeper models, larger patch sizes, overlapped image patches, label smoothing~\cite{muller2019does}, and stochastic depth~\cite{huang2016deep}. Previous research has shown that some classic models can achieve state-of-the-art performance through carefully-designed training regimes~\cite{strikesback},
    \item We tried a number of ways to split and mix the channels and learned that some perform better than the others. There might be even better approaches to do this,
    \item Incorporating techniques similar to the ones proposed here to optimize other MLP-like models is also a promising direction,
    \item MLP-like models, including SplitMixer, lack effective means of explanation and visualization, which need to be addressed in the future (See Appendix~\ref{appx:maps}),
    \item Our results entertain the idea that it may be possible to find model classes that have fewer parameters than the number of data points. This may challenge the current belief that deep networks have to be overparameterized to perform well,
    \item Large internal resolution and isotropic design make SplitMixer-type models appealing for vision tasks such as semantic segmentation and object detection. This direction needs more exploration. Some works have already reported promising results regarding this (\eg~\cite{permutator,lian2021mlp,chen2021cyclemlp}), and
    \item Lastly, it is essential to a) understand the basic principles underlying CNNs, Transformers, and MLP-based models, b) explain why some design choices work while some others do not (\eg~\cite{park2022vision}), and c) enumerate the key must-have building blocks (\eg convolution, pooling, residual connection, normalization, data augmentation, and patch embeddings). This will help unify existing architectures, improve them, and invent even more efficient ones.
\end{itemize}

\noindent {\bf Note: We are currently running the experiments on ImageNet-1K dataset.}

\bibliography{references}
\bibliographystyle{plain}

\newpage
\clearpage

\appendix

\section{Parameter and FLOPS Saving}
\label{appx:params}

Here, we empirically show how much parameters and FLOPS can be reduced by SplitMixer. Experiments are conducted on CIFAR-10 and ImageNet datasets using model specifications mentioned in the main text. 

\begin{figure}[htbp]
    \vspace{20pt}
    \centering
    \includegraphics[width=5cm,height=8.6cm]{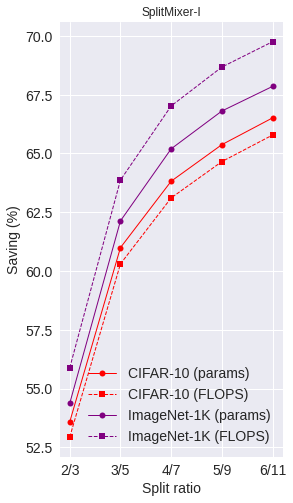} 
    \includegraphics[width=5cm,height=8.6cm]{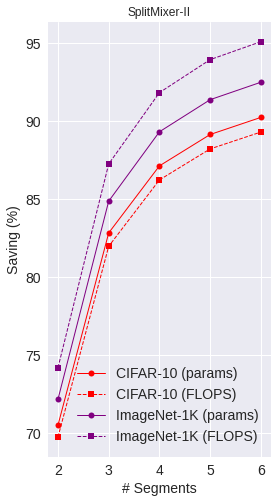} \\
    \includegraphics[width=5cm,height=8.6cm]{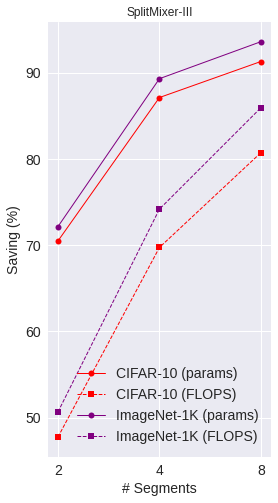} 
    \includegraphics[width=5cm,height=8.6cm]{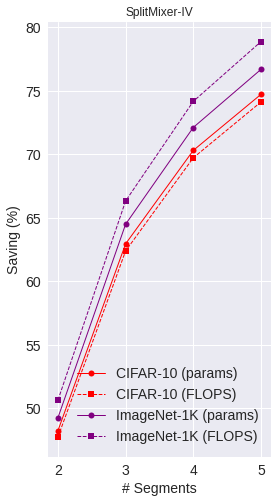} \\    
    \caption{Potential savings in parameters and FLOPS for different SplitMixer variants.}
    \label{fig:Savings}
\end{figure}

\newpage
\section{PyTorch Implementations}
\label{appx:implementations}

\begin{figure}[htbp]
{\footnotesize
\begin{minted}[linenos]{python}

class ChannelMixerI(nn.Module):
    ''' Partial overlap; In each block only one segment is convolved. '''
    def __init__(self, hdim, is_odd=0, ratio=2/3, **kwargs):
        super().__init__()
        self.hdim = hdim
        self.partial_c = int(hdim *ratio)
        self.mixer = nn.Conv2d(self.partial_c, self.partial_c, kernel_size=1)
        self.is_odd = is_odd
    
    def forward(self, x):
        if self.is_odd == 0:
            idx = self.partial_c
            return torch.cat((self.mixer(x[:, :idx]), x[:, idx:]), dim=1)
        else:
            idx = self.hdim - self.partial_c
            return torch.cat((x[:, :idx], self.mixer(x[:, idx:])), dim=1)


def SplitMixerI(dim, blocks, kernel_size=5, patch_size=2, n_classes=10, ratio=2/3):
    return nn.Sequential(
        nn.Conv2d(3, dim, kernel_size=patch_size, stride=patch_size),
        nn.GELU(),
        nn.BatchNorm2d(dim),
        *[nn.Sequential(
                Residual(nn.Sequential(
                    nn.Conv2d(dim, dim, (1,kernel_size), groups=dim, padding="same"),
                    nn.GELU(),
                    nn.BatchNorm2d(dim),
                    nn.Conv2d(dim, dim, (kernel_size,1), groups=dim, padding="same"),
                    nn.GELU(),
                    nn.BatchNorm2d(dim)
                )),
                ChannelMixerI(dim, i % 2,ratio),
                nn.GELU(),
                nn.BatchNorm2d(dim),
        ) for i in range(blocks)],
        nn.AdaptiveAvgPool2d((1,1)),
        nn.Flatten(),
        nn.Linear(dim, n_classes)
    )
    
    
def SplitMixerI_channel_only(dim, blocks, kernel_size=5, patch_size=2, n_classes=10, ratio=2/3):
    return nn.Sequential(
        nn.Conv2d(3, dim, kernel_size=patch_size, stride=patch_size),
        nn.GELU(),
        nn.BatchNorm2d(dim),
        *[nn.Sequential(
                Residual(nn.Sequential(
                    nn.Conv2d(dim, dim, (kernel_size,kernel_size), groups=dim, padding="same"),
                    nn.GELU(),
                    nn.BatchNorm2d(dim),
                )),
                ChannelMixerI(dim, i % 2,ratio),
                nn.GELU(),
                nn.BatchNorm2d(dim),
        ) for i in range(blocks)],
        nn.AdaptiveAvgPool2d((1,1)),
        nn.Flatten(),
        nn.Linear(dim, n_classes)
    )

\end{minted}
}
    \caption{SplitMixer-I.}
    \label{SplitMixerI}
    \vspace*{-8em}  
\end{figure}

\begin{figure}[htbp]
{\footnotesize
\begin{minted}[linenos]{python}


class ChannelMixerII(nn.Module):
    ''' No overlap; In each block only one segment is convolved. '''
    def __init__(self, hdim, remainder=0, num_segments=3, **kwargs):
        super().__init__()
        self.hdim = hdim
        self.remainder = remainder
        self.num_segments = num_segments
        self.bin_dim = int(hdim / num_segments)
        self.c = hdim - self.bin_dim * (num_segments - 1) if (
            remainder == num_segments - 1) else self.bin_dim
        self.mixer = nn.Conv2d(self.c, self.c, kernel_size=1)
    
    def forward(self, x):
        start = self.remainder * self.bin_dim
        end = self.hdim if (self.remainder == self.num_segments - 1) else (
            (self.remainder + 1) * self.bin_dim)
        return torch.cat((x[:, :start], self.mixer(x[:, start : end]),
                          x[:, end:]), dim=1)


def SplitMixerII(dim, blocks, kernel_size=5, patch_size=2, n_classes=10, n_part=2):
    return nn.Sequential(
        nn.Conv2d(3, dim, kernel_size=patch_size, stride=patch_size),
        nn.GELU(),
        nn.BatchNorm2d(dim),
        *[nn.Sequential(
                Residual(nn.Sequential(
                    nn.Conv2d(dim, dim, (1,kernel_size), groups=dim, padding="same"),
                    nn.GELU(),
                    nn.BatchNorm2d(dim),
                    nn.Conv2d(dim, dim, (kernel_size,1), groups=dim, padding="same"),
                    nn.GELU(),
                    nn.BatchNorm2d(dim)
                )),
                ChannelMixerII(dim, i % n_part, n_part),
                nn.GELU(),
                nn.BatchNorm2d(dim),
        ) for i in range(blocks)],
        nn.AdaptiveAvgPool2d((1,1)),
        nn.Flatten(),
        nn.Linear(dim, n_classes)
    )


\end{minted}
}
    \caption{SplitMixer-II.}
    \label{SplitMixerII}

\end{figure}

\begin{figure}[htbp]
{\footnotesize
\begin{minted}[linenos]{python}

class ChannelMixerIII(nn.Module):
    ''' No overlap; In each block all segments are convolved;
        Parameters are shared across segments. '''
    def __init__(self, hdim, num_segments=3, **kwargs):
        super().__init__()
        assert hdim % num_segments == 0, (
            f'hdim {hdim} need to be divisible by num_segments {num_segments}')
        self.hdim = hdim
        self.num_segments = num_segments
        self.c = hdim // num_segments
        self.mixer = nn.Conv2d(self.c, self.c, kernel_size=1)
    
    def forward(self, x):
        c = self.c
        x = [self.mixer(x[:, c * i : c * (i + 1)]) for i in range(self.num_segments)]
        return torch.cat(x, dim=1)


def SplitMixerIII(dim, blocks, kernel_size=5, patch_size=2, n_classes=10, n_part=2):
    return nn.Sequential(
        nn.Conv2d(3, dim, kernel_size=patch_size, stride=patch_size),
        nn.GELU(),
        nn.BatchNorm2d(dim),
        *[nn.Sequential(
                Residual(nn.Sequential(
                    nn.Conv2d(dim, dim, (1,kernel_size), groups=dim, padding="same"),
                    nn.GELU(),
                    nn.BatchNorm2d(dim),
                    nn.Conv2d(dim, dim, (kernel_size,1), groups=dim, padding="same"),
                    nn.GELU(),
                    nn.BatchNorm2d(dim)
                )),
                ChannelMixerIII(dim, n_part),
                nn.GELU(),
                nn.BatchNorm2d(dim),
        ) for i in range(blocks)],
        nn.AdaptiveAvgPool2d((1,1)),
        nn.Flatten(),
        nn.Linear(dim, n_classes)
    )
\end{minted}
}
    \caption{SplitMixer-III.}
    \label{SplitMixerIII}
\end{figure}

\begin{figure}[h]
{\footnotesize
\begin{minted}[linenos]{python}

class ChannelMixerIV(nn.Module):
    ''' No overlap; In each block all segments are convolved;
      No parameter sharing across segments. '''
    def __init__(self, hdim, num_segments=3, **kwargs):
        super().__init__()
        self.hdim = hdim
        self.num_segments = num_segments
        c = hdim // num_segments
        last_c = hdim - c * (num_segments - 1)
        self.mixer = nn.ModuleList(
            [nn.Conv2d(c, c, kernel_size=1) for _ in range(num_segments - 1)
            ] + ([nn.Conv2d(last_c, last_c, kernel_size=1)]))
        self.c, self.last_c = c, last_c
   
    def forward(self, x):
        c, last_c = self.c, self.last_c
        x = [self.mixer[i](x[:, c * i : c * (i + 1)]) for i in (
            range(self.num_segments - 1))] + [self.mixer[-1](x[:, -last_c:])]
        return torch.cat(x, dim=1)


def SplitMixerIV(dim, blocks, kernel_size=5, patch_size=2, n_classes=10, n_part=2):
    return nn.Sequential(
        nn.Conv2d(3, dim, kernel_size=patch_size, stride=patch_size),
        nn.GELU(),
        nn.BatchNorm2d(dim),
        *[nn.Sequential(
                Residual(nn.Sequential(
                    nn.Conv2d(dim, dim, (1,kernel_size), groups=dim, padding="same"),
                    nn.GELU(),
                    nn.BatchNorm2d(dim),
                    nn.Conv2d(dim, dim, (kernel_size,1), groups=dim, padding="same"),
                    nn.GELU(),
                    nn.BatchNorm2d(dim)
                )),
                ChannelMixerIV(dim, n_part),
                nn.GELU(),
                nn.BatchNorm2d(dim),
        ) for i in range(blocks)],
        nn.AdaptiveAvgPool2d((1,1)),
        nn.Flatten(),
        nn.Linear(dim, n_classes)
    )

\end{minted}
}
    \caption{SplitMixer-IV.}
    \label{SplitMixerIV}

\end{figure}

\clearpage

\newpage
\section{Model Throughput}
\label{appx:throughput}

\begin{table}[h]
\begin{small}
\begin{center}
\begin{tabularx}{.8\textwidth}{l|YYYYYYY}
\toprule
Network& \multicolumn{7}{c}{Throughput \ \ (img/sec)} \\
\hline \hline
ConvMixer & \multicolumn{7}{c}{483.37} \\ \hline 
 & \multicolumn{7}{c}{Overlap ratio} \\ \cline{2-8}
         & 2/3 & 3/5 & 4/7 & 5/9 & 6/11 & - & -  \\ \cline{2-8}
SplitMixer-I & 733.57 & 742.80 & 746.88 & 734.04 & 734.07 &  - &  -  \\ \hline 
 & \multicolumn{7}{c}{Number of segments} \\ \cline{2-8} 
                    & 2 & 3 & 4 & 5 & 6 & 7 & 8 \\ \cline{2-8}
SplitMixer-II      & 770.18 &  790.13 & 815.27 & 821.58 & 825.91 &  - & -  \\
SplitMixer-III       & 699.80 & - &  718.14 & - & - & - & 718.25  \\
SplitMixer-IV      &  696.39 & 681.12 & 715.64 & 714.95 & - & - & - \\
\bottomrule
\end{tabularx}
\end{center}
\caption{Model throughput for SplitMixer-A-256/8 on a Tesla v100 GPU with 16GB RAM over a batch of 64 images of size 224 $\times$ 224, averaged over 100 such batches.}
\label{tab:throughput-appx}
\end{small}
\end{table}

\clearpage

\newpage
\section{Other Variants of SplitMixer}
\label{appx:SplitMixerV}

Here, we introduce another variant of SplitMixer, called SplitMixer-V, which is similar to SplitMixer-I with the difference that in each block both overlapped segmented are updated. The approximate reduction in parameters per block is:
\begin{equation}
    h^2 - 2 \times (\alpha \times h)^2  = (1 - 2 \times \alpha^2)  \times h^2 
\end{equation} 
\noindent which means $1 - 2\alpha^2$ fraction of parameters are reduced (\eg 11\% parameter reduction for $\alpha=2/3$, much lower than 56\% obtained using SplitMixer-I with the same $\alpha$). The saving in FLOPS is the same. 

We tested SplitMixer-V over CIFAR-\{10,100\} datasets with the same parameters used for other SplitMixers on these datasets. Results are shown in Table below. This model performs close to ConvMixer and is sometimes even better than it. It, however, does not save much of the parameters and FLOPS, compared to the other SplitMixers. As was mentioned in the main text, SplitMixer variants offer different degrees of trade-off between accuracy and the number of parameters (and FLOPS).

\begin{table}[h]
\begin{small}
\begin{center}
\begin{tabularx}{.9\textwidth}{l|YYY|YYY}
\cline{2-7}
     & \multicolumn{3}{c|}{CIFAR-10} & \multicolumn{3}{c}{CIFAR-100} \\
     \cline{2-7}
     &  Accuracy (\%) & Params (M) & FLOPS (M) &  Accuracy (\%) & Params (M) & FLOPS (M) \\
\hline \hline
ConvMixer & 94.17 & 0.59 & 152 & 73.92 &  0.62 & 152 \\ \bottomrule
SplitMixer-V-256/8 & & & & & & \\
 $\ \ \ \ \ \ \ \ \ \ \ \ \ \ \ \ \alpha = 2/3$ & 93.88 & 0.51 & 131 &     73.68 & 0.53 & 131 \\ \cline{2-7}
$\ \ \ \ \ \ \ \ \ \ \ \ \ \ \ \ \alpha = 3/5$ & 93.96 & 0.42 & 108    &  74.63 & 0.44 & 108\\
\bottomrule
\end{tabularx}
\end{center}
\caption{Performance of the SplitMixer-V-256/8 model \emph{vs.} ConvMixer.}
\label{tab:newmixer}
\end{small}
\end{table}

The code for this design is given below.

\begin{figure}[h]
{\footnotesize
\begin{minted}[linenos]{python}

class ChannelMixerV(nn.Module):
    ''' Partial overlap; In each block all segments are convolved;
      No parameter sharing across segments. '''
    def __init__(self, hdim, ratio=2/3, **kwargs):
        super().__init__()
        self.hdim = hdim
        self.c = int(hdim *ratio)
        self.mixer1 = nn.Conv2d(self.c, self.c, kernel_size=1)
        self.mixer2 = nn.Conv2d(self.c, self.c, kernel_size=1)
    
    def forward(self, x):
        c, hdim = self.c, self.hdim
        x = torch.cat((self.mixer1(x[:, :c]), x[:, c:]), dim=1)
        return torch.cat((x[:, :(hdim - c)], self.mixer2(x[:, (hdim - c):])), dim=1)


def SplitMixer-V(dim, blocks, kernel_size=5, patch_size=2, n_classes=10, ratio=2/3):
    return nn.Sequential(
        nn.Conv2d(3, dim, kernel_size=patch_size, stride=patch_size),
        nn.GELU(),
        nn.BatchNorm2d(dim),
        *[nn.Sequential(
                Residual(nn.Sequential(
                    nn.Conv2d(dim, dim, (1,kernel_size), groups=dim, padding="same"),
                    nn.GELU(),
                    nn.BatchNorm2d(dim),
                    nn.Conv2d(dim, dim, (kernel_size,1), groups=dim, padding="same"),
                    nn.GELU(),
                    nn.BatchNorm2d(dim)
                )),
                ChannelMixerV(dim, ratio),
                nn.GELU(),
                nn.BatchNorm2d(dim),
        ) for i in range(blocks)],
        nn.AdaptiveAvgPool2d((1,1)),
        nn.Flatten(),
        nn.Linear(dim, n_classes)
    )

\end{minted}
}
    \vspace*{-1em}
    \caption{SplitMixer-V.}
    \label{SplitMixerV}

\end{figure}

\clearpage

\newpage
\section{Channel Mixing Using 3D Convolution}
\label{appx:3dConv}

\begin{figure}[h]
{\footnotesize
\begin{minted}[linenos]{python}

class StrideMixer(nn.Module):
    def __init__(self, dim, blocks, kernel_size=5, patch_size=2, num_classes=10,
                 channel_kernel_size=128, channel_stride=128): # setting III
        super().__init__()
        self.patch_emb = nn.Sequential(
            nn.Conv2d(3, dim, kernel_size=patch_size, stride=patch_size),
            nn.GELU(),
            nn.BatchNorm2d(dim),)

        # calculate channel conv h_out
        c_out = (dim - channel_kernel_size) / channel_stride + 1
        assert dim % c_out == 0, 'setting is not valid, double check channel_kernel_size and channel_stride'
        h_out = int(dim / c_out)

        self.spatial_mixer, self.channel_mixer = ModuleList(), ModuleList()
        for _ in range(blocks):
            spatial_mixer = Residual(nn.Sequential(
                nn.Conv2d(dim, dim, (1, kernel_size), groups=dim, padding="same"),
                nn.GELU(),
                nn.BatchNorm2d(dim),
                nn.Conv2d(dim, dim, (kernel_size, 1), groups=dim, padding="same"),
                nn.GELU(),
                nn.BatchNorm2d(dim)))
            self.spatial_mixer.append(spatial_mixer)

            channel_mixer = nn.Sequential(
                nn.Conv3d(1, h_out, (channel_kernel_size, 1, 1), stride=(channel_stride, 1, 1)),
                nn.GELU(),
                nn.Flatten(1, 2),
                nn.BatchNorm2d(dim))
            self.channel_mixer.append(channel_mixer)

        self.head = nn.Sequential(
            nn.AdaptiveAvgPool2d((1,1)),
            nn.Flatten(),
            nn.Linear(dim, num_classes)
            )

        self.blocks = blocks

    def forward(self, x):
        x = self.patch_emb(x)
        for i in range(self.blocks):
            x = self.spatial_mixer[i](x).unsqueeze(1)
            x = self.channel_mixer[i](x)
        return self.head(x)

\end{minted}
}
    \caption{Channel mixing using 3D convolution.}
    \label{SplitMixer3D}

\end{figure}

\clearpage

\newpage
\section{Visualization of Feature Maps}
\label{appx:maps}

Here, we visualize the output maps of the SplitMixer-I-256/8 model trained over CIFAR-10 with $\alpha=2/3$, $p=2$, and $k=5$. Notice that in each block 170 channels are convolved (first 170 channels in even blocks and the second 170 channels in odd blocks; with overlap). Notice that the learned $2 \times 2$ kernels for patch embedding are too small to plot here. 

\begin{figure}[htbp]
    \centering
    \includegraphics[width=.3\textwidth]{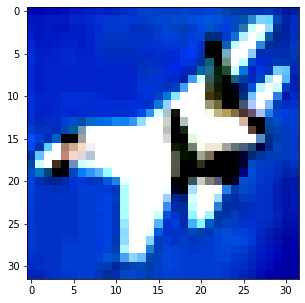} \\
    \vspace{10pt}
    \includegraphics[width=\textwidth]{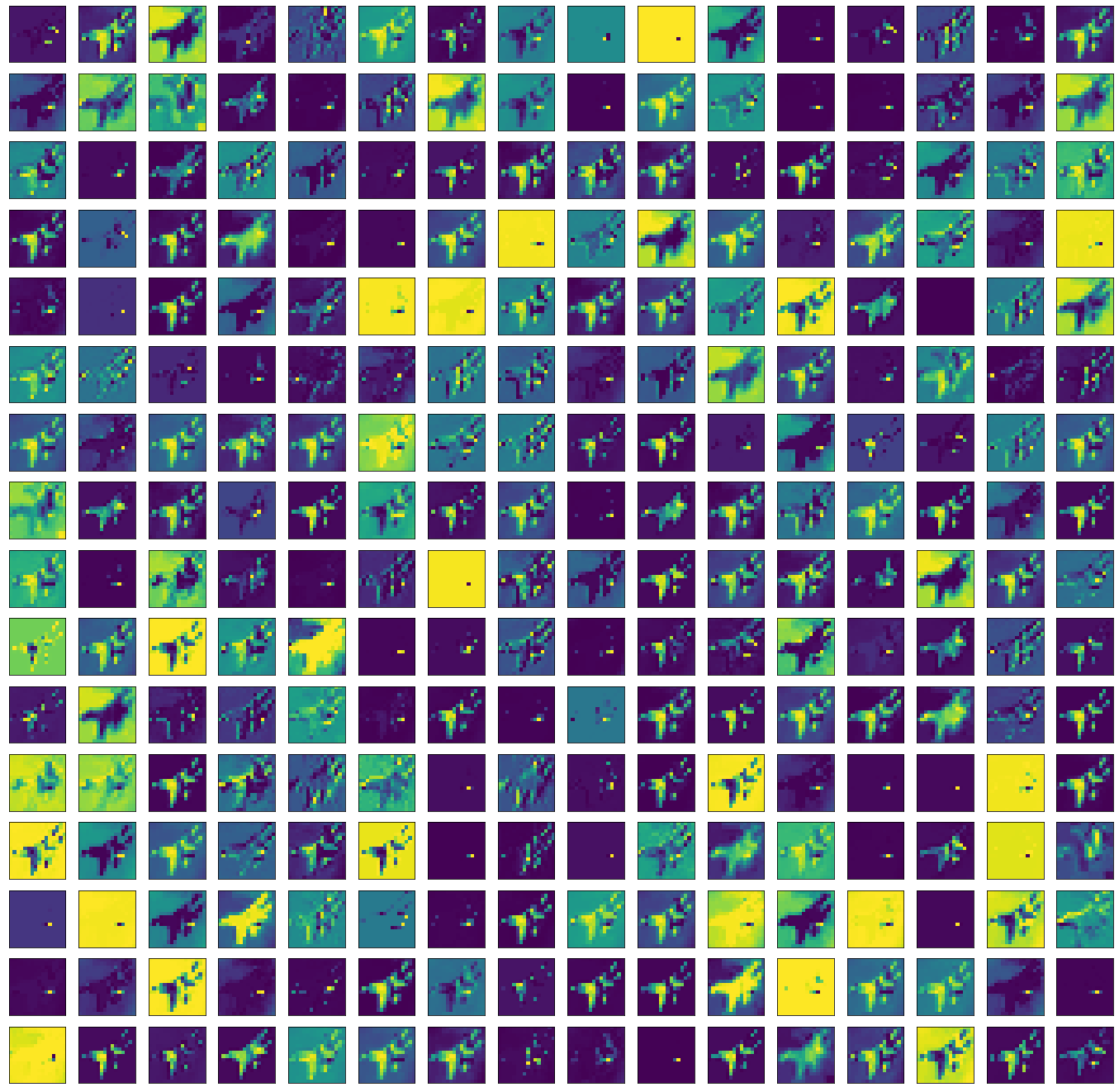} 
    \caption{Top) A sample image from CIFAR-10 dataset, Bottom) The 256 feature maps after patch embedding (\ie after the first conv2d).}
    \label{fig:vis0}
    \vspace{-200pt}
\end{figure}

\begin{figure}[htbp]
    \centering
    \includegraphics[width=1\textwidth]{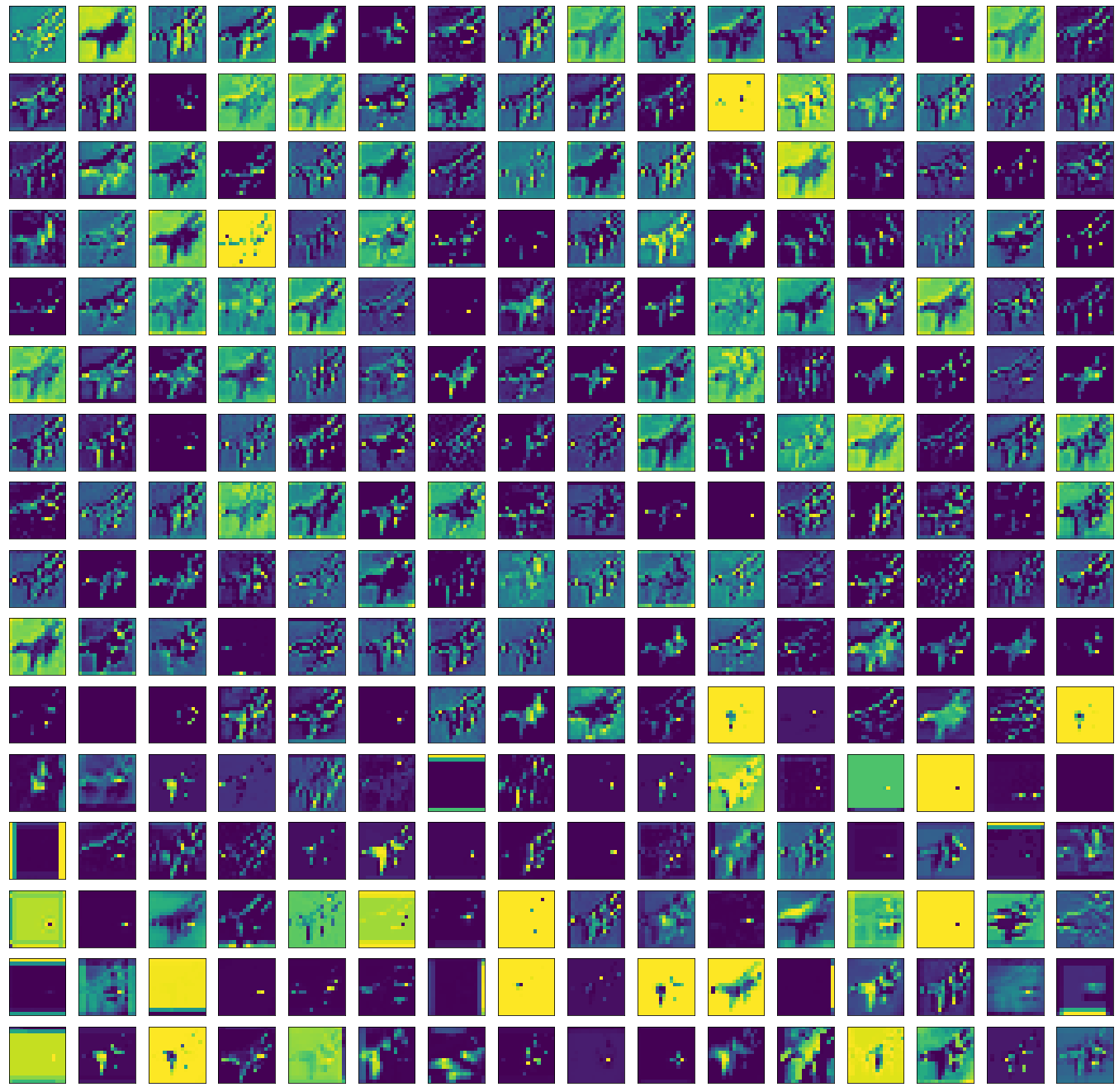} \\
    \caption{The output maps after the first SplitMixer block.}
    \label{fig:vis1}
\end{figure}

\begin{figure}[htbp]
    \centering
    \includegraphics[width=1\textwidth]{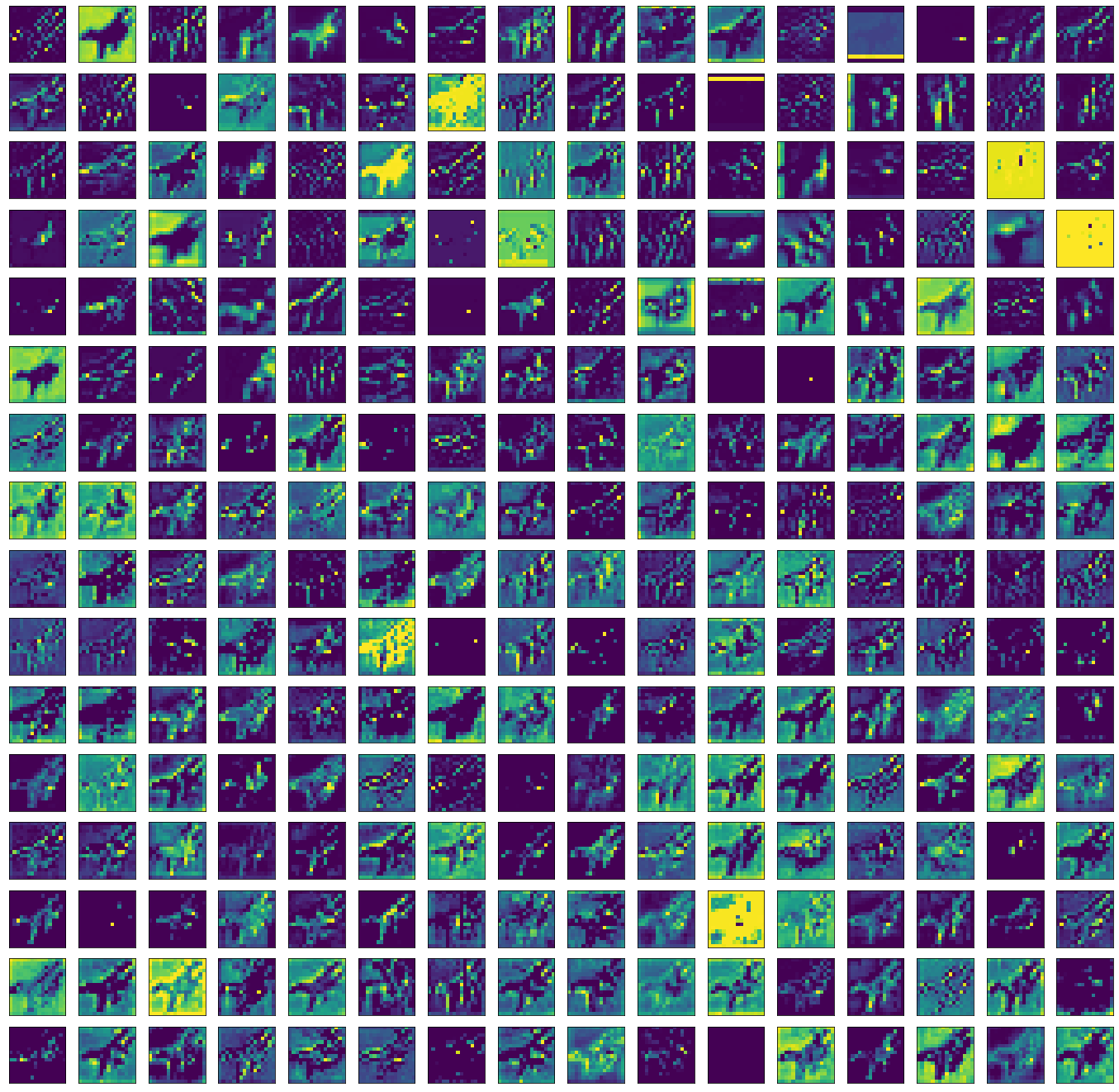} \\
    \caption{The output maps after the second SplitMixer block.}
    \label{fig:vis2}
\end{figure}

\begin{figure}[htbp]
    \centering
    \includegraphics[width=1\textwidth]{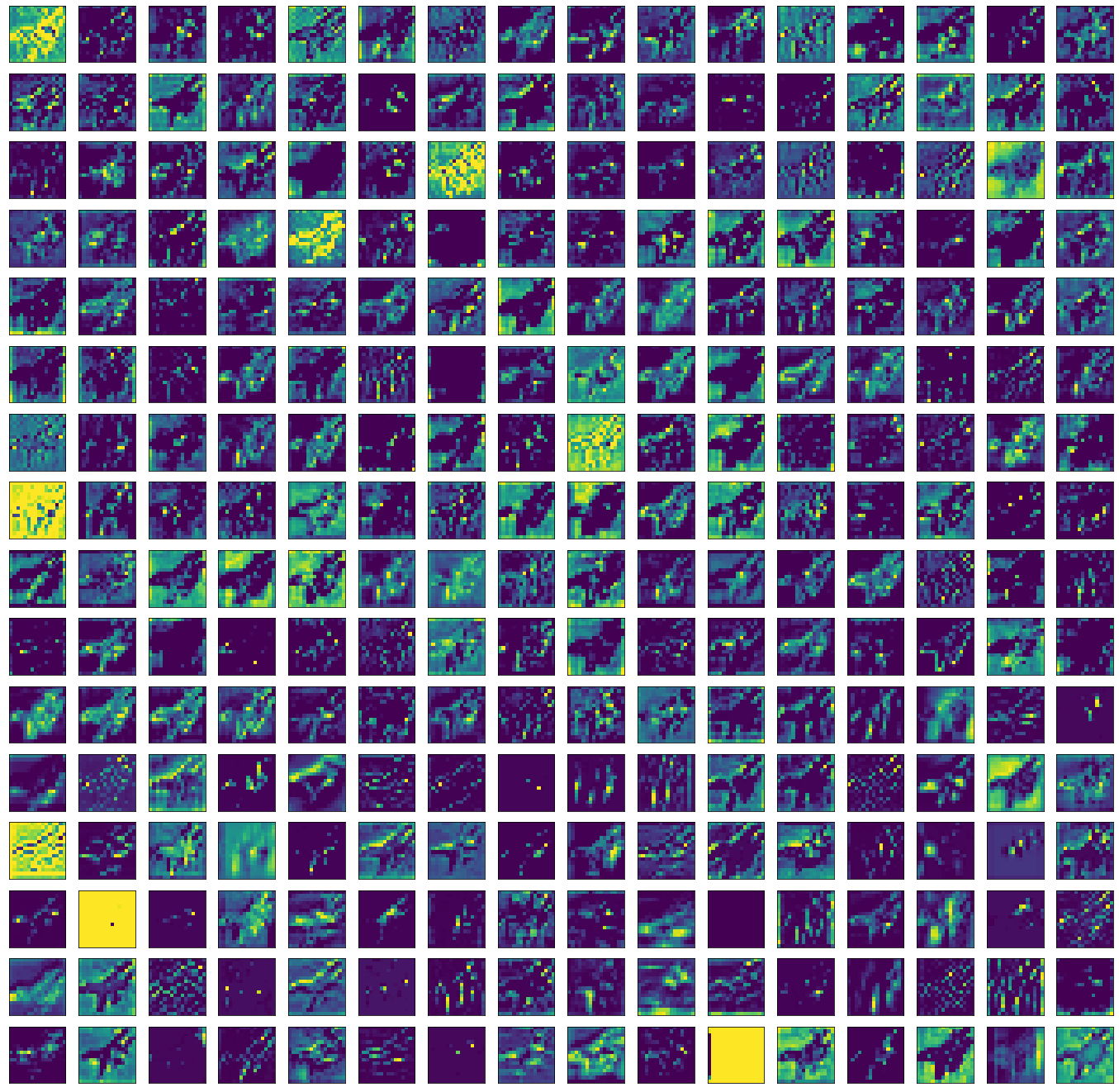} \\
    \caption{The output maps after the third SplitMixer block.}
    \label{fig:vis3}
\end{figure}

\begin{figure}[htbp]
    \centering
    \includegraphics[width=1\textwidth]{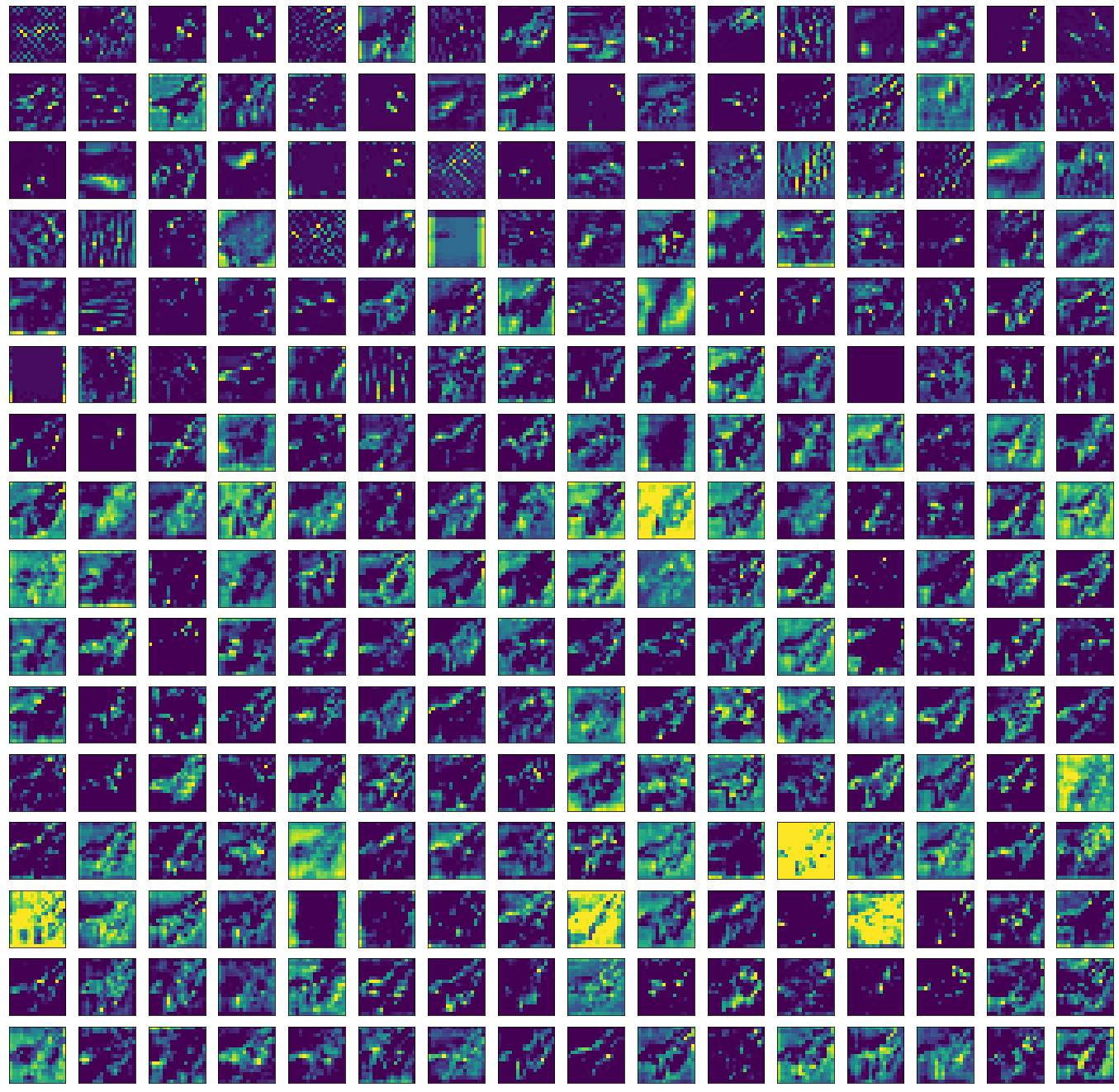} \\
    \caption{The output maps after the fourth SplitMixer block.}
    \label{fig:vis4}
\end{figure}

\end{document}